\documentclass[sigconf,nonacm]{acmart}

\AtBeginDocument{%
  }

\usepackage{comment}
\usepackage{hhline}
\usepackage[linewidth=1pt]{mdframed}
\usepackage{multirow}
\usepackage{xspace}

\usepackage[acronym, nowarn]{glossaries}
\newacronym{ai}{AI}{Artificial Intelligence}
\newacronym{bleu}{BLEU}{Bilingual Evaluation Understudy}
\newacronym{cider}{CIDEr}{Consensus-based Image Description Evaluation}
\newacronym{cnn}{CNN}{Convolutional Neural Network}
\newacronym{fid}{FID}{Fréchet Inception Distance}
\newacronym{gi}{GI}{Gastrointestinal}
\newacronym{task-is}{IS}{Image Synthesis}
\newacronym{is}{IS}{Inception Score}
\newacronym{lora}{LoRA}{Low-Rank Adaptation}
\newacronym{meteor}{METEOR}{Metric for Evaluation of Translation with Explicit ORdering}
\newacronym{ml}{ML}{Machine Learning}
\newacronym{mmdit}{MMDiT}{Multimodal Diffusion Transformer}
\newacronym{nlp}{NLP}{Natural Language Processing}
\newacronym{task-opg}{OPG}{Optimal Prompt Generation}
\newacronym{vqa}{VQA}{Visual Question Answering}

\newacronym{nn}{NN}{Neural Network}
\newacronym{tp}{TP}{True Positive}
\newacronym{tn}{TN}{True Negative}
\newacronym{fn}{FN}{False Negative}
\newacronym{fp}{FP}{False Positive}
\newacronym{rnn}{RNN}{Recurrent Neural Network}
\newacronym{vit}{ViT}{Vision Transformer}
\newacronym{lstm}{LSTM}{Long Short-Term Memory}
\newacronym{fnn}{FNN}{Feedforward Neural Network}
\newacronym{3d-cnn}{3D-CNN}{3D Convolutional Neural Network}
\newacronym{2d-cnn}{2D-CNN}{2D Convolutional Neural Network}
\newacronym{cv}{CV}{Computer Vision}

\copyrightyear{2024} 
\acmYear{2024} 
\setcopyright{acmlicensed}
\acmConference[VLM4Bio '24]{Proceedings of the
First International Workshop on Vision-Language Models for Biomedical
Applications}{October 28-November 1 2024}{Melbourne, VIC, Australia}
\acmBooktitle{Proceedings of the First International Workshop on
Vision-Language Models for Biomedical Applications (VLM4Bio '24), October
28-November 1 2024, Melbourne, VIC, Australia}
\acmDOI{10.1145/3689096.3689458}
\acmISBN{979-8-4007-1207-4/24/10}

\newcommand{\dsname}{Kvasir-VQA\xspace}

\begin{document}

\title{\dsname: A Text-Image Pair GI Tract Dataset}

\settopmatter{authorsperrow=4}

\author{Sushant Gautam}
\affiliation{%
 \institution{SimulaMet}
 \country{Norway}
}
\authornote{Also affiliated with Oslo Metropolitan University, Norway.}

\author{Andrea Storås}
\affiliation{%
 \institution{SimulaMet}
 \country{Norway}
}

\author{Cise Midoglu}
\affiliation{%
 \institution{SimulaMet}
 \country{Norway}
 }

\author{Steven A. Hicks}
\affiliation{%
 \institution{SimulaMet}
 \country{Norway}
}

\author{Vajira Thambawita}
\affiliation{%
 \institution{SimulaMet}
 \country{Norway}
}

\author{Pål Halvorsen}
\affiliation{%
 \institution{SimulaMet}
 \country{Norway}
 }
\authornotemark[1]

\author{Michael A. Riegler}
\affiliation{%
\institution{SimulaMet}
\country{Norway}
}
\authornotemark[1]

\renewcommand{\shortauthors}{Sushant Gautam et al.}

\begin{abstract}
We introduce \dsname, an extended dataset derived from the HyperKvasir and Kvasir-Instrument datasets, augmented with question-and-answer annotations to facilitate advanced machine learning tasks in \gls{gi} diagnostics. This dataset comprises 6,500 annotated images spanning various \gls{gi} tract conditions and surgical instruments, and it supports multiple question types including yes/no, choice, location, and numerical count. The dataset is intended for applications such as image captioning, \gls{vqa}, text-based generation of synthetic medical images, object detection, and classification. Our experiments demonstrate the dataset's effectiveness in training models for three selected tasks, showcasing significant applications in medical image analysis and diagnostics. We also present evaluation metrics for each task, highlighting the usability and versatility of our dataset.
The dataset and supporting artifacts are available at \\\href{https://datasets.simula.no/kvasir-vqa}{https://datasets.simula.no/kvasir-vqa}.
\end{abstract}



\begin{teaserfigure}\includegraphics[width=.8\textwidth]{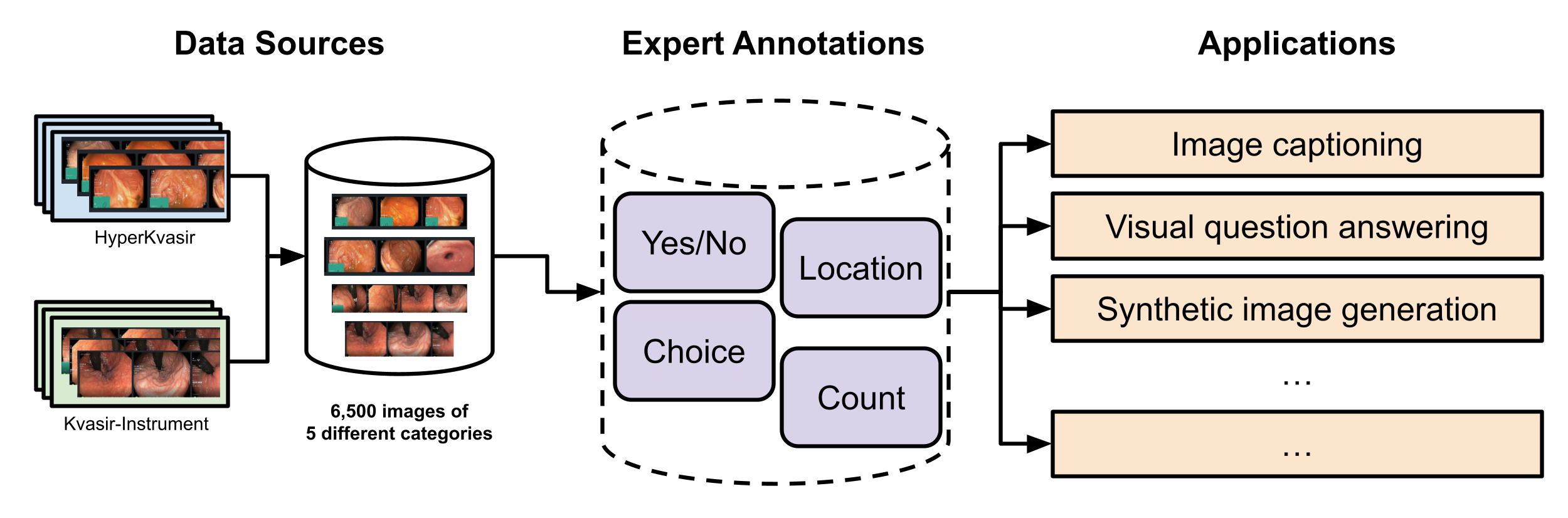}
    \centering
    \caption{Overview of the sources, curation, and applications of the \dsname dataset.}
    \Description[Brief overview of dataset sources and applications]{A diagram illustrating the sources of the \dsname dataset, the curation process, and various applications of the dataset. The diagram includes labeled sections showing data sources, data handling steps, and different ways the dataset is used.}
    \label{fig:teaser}
\end{teaserfigure}

\maketitle

\section{Introduction}\label{sec:introduction}

The advancement of medical diagnostics increasingly relies on the integration of \gls{ai} and \gls{ml} techniques to analyze complex medical images. The human digestive system is categorized into the upper and lower \acrfull{gi} tracts and includes the esophagus, stomach, and intestines. Diseases in the \gls{gi} tract, being prevalent and often requiring intricate diagnostic procedures, present an ideal domain for deploying AI-driven solutions. Existing datasets such as HyperKvasir~\cite{Borgli2020Aug} and Kvasir-Instrument~\cite{Jha2021Jan} have made substantial contributions to this field by offering a wide range of classification- and segmentation-labeled \gls{gi} images. 

However, the lack of comprehensive textual annotations, particularly question-and-answer pairs, limits the potential for developing advanced \gls{ai} models capable of nuanced understanding and decision-making. This limitation is critical because question-and-answer pairs for an input can help to simulate the reasoning process that we humans go through, thereby enabling \gls{ai} models to better understand context, interpret complex scenarios, and provide more accurate diagnostics. Furthermore, these annotations can facilitate the training of \gls{ai} systems to handle a wider variety of cases, improving their generalizability and reliability in real-world clinical settings.

\newpage

In this study, we extend these foundational datasets by incorporating detailed question-and-answer annotations (Figure~\ref{fig:teaser}). This enhancement aims to facilitate a broader range of \gls{ml} applications, such as image captioning, \acrfull{vqa}, text-based synthetic image generation, object detection, and classification. By curating a new dataset, \dsname, we bridge the gap between medical image analysis and practical diagnostic tools, ultimately aiming to improve patient outcomes and diagnostic precision. We present preliminary experiments with \dsname in order to demonstrate three of these use cases: image captioning, \gls{vqa}, and synthetic medical
image generation. These tasks were selected to highlight different aspects of the dataset and its potential use in various contexts.

The rest of this paper is structured as follows. Section~\ref{sec:background} provides a comprehensive background on existing medical image analysis datasets, particularly HyperKvasir and Kvasir-Instrument, and their contributions to the field. Section~\ref{sec:dataset} details our curation of the \dsname dataset, including the incorporation of question-and-answer annotations, and provides an elaborate dataset description, specifying the image categories and types of questions included. Section~\ref{sec:experiments} presents the experiments we have conducted to demonstrate the dataset’s utility, namely image captioning, \gls{vqa}, and synthetic medical image generation. In Section~\ref{sec:discussion}, we summarize the key findings, discuss the potential applications of our dataset in medical diagnostics as well as its limitations, and outline directions for future research. Section~\ref{sec:conclusion} concludes the paper. 

\section{Background and Related Work}\label{sec:background}

The human gastrointestinal (GI) tract is susceptible to a wide range of abnormal mucosal conditions, varying from minor irritations to highly lethal diseases~\cite{Navarre2002,Gelberg2018}. According to the International Agency for Research on Cancer, a specialized cancer agency of the World Health Organization (WHO), \gls{gi} cancers account for approximately 4.8 million new cases annually worldwide~\cite{Arnold2020Jul}. These cancers often have a high mortality rate, contributing to around 3.4 million deaths each year~\cite{Arnold2020Jul}.

Endoscopy is the current gold-standard procedure for examining the \gls{gi} tract, but its effectiveness is limited by the operator's performance, which results in a significant average miss rate of around 20\% for polyps in the colon~\cite{Ahmad2019Jan,Wang2019Oct}. Consequently, improving endoscopic performance, enhancing the quality of clinical examinations, and implementing systematic screening are crucial for reducing morbidity and mortality associated with \gls{gi} diseases~\cite{Kaminski2017Apr,Saftoiu2020Apr}.

The emergence of AI-enabled support systems offers promise in assisting healthcare professionals in providing high-quality care on a large scale~\cite{Alowais2023Dec}. These \gls{ai} systems, particularly those using \gls{ml} techniques, require extensive training on well-curated datasets containing human-verified annotations to be effective in real-world tasks, such as detecting precancerous lesions or cancers in medical images~\cite{Koh2022Oct}. The performance of \gls{ai} in medical image analysis has seen significant advancements, primarily driven by the quality of the datasets and the sophistication of the algorithms employed~\cite{Panayides2020May,Altaf2019Jul}.

\subsection{Gastrointestinal Image Datasets}

Earlier \gls{gi} datasets have focused on different findings such as polyps~\cite{dataset2017-polyp,bernal2015wm,bernal2012towards,tajbakhsh2015automated,silva2014toward,gianadataset2017,angermann2017towards,bernal2018polyp,kvasir2017mmsys,jha2020kvasir}, 
endoscopic artifacts~\cite{ali2019endoscopy}, \gls{gi} lesions~\cite{gastrolab,wco,wco1,atlas,gastroatlas}, angiectasia, bleeding, inflammation, esophagitis, ulcerative colitis, Z/line, pylorus, cecum, dyed resection margins, and 
stool~\cite{Koulaouzidis2017,gianadataset2017,angermann2017towards,bernal2018polyp,kvasir2017mmsys,nerthus2017mmsys}. %
However, a number of these datasets are not publicly available, serve more as educational databases rather than being suitable for algorithm training, and/or are not usable for \gls{ml}. Several other datasets have also focused specifically on diagnostic and therapeutic tool segmentation in endoscopy~\cite{bodenstedt2018comparative,allan20192017,allan20192018,ross2020robust,Jha2021Jan}.

Kvasir~\cite{Pogorelov2017Kvasir}, HyperKvasir~\cite{Borgli2020Aug}, Kvasir-SEG~\cite{jha2020kvasir}, Kvasir-Capsule~\cite{Smedsrud2021May} and Kvasir-Instrument~\cite{Jha2021Jan} are prominent datasets that have catalyzed research in \gls{gi} diagnostics, see for example~\cite{Jha2021KvasirReviewPerformance,Tzavara2021KvasirInstrumentTransfer,Fagereng2022PolypConnectHyperkvasir,Wu2021HyperkvasirForHAL,Srivastava2022}, providing a large number of \gls{gi} images of various types and some with bounding boxes and segmentation masks. 
Recent studies have leveraged these datasets to develop models for specific tasks, such as polyp detection using \glspl{cnn} and capsule endoscopy analysis. However, these models are limited by the scope of available annotations, which primarily focus on image classification, object detection, and segmentation.

\subsection{Image Captioning}
Image captioning in medical imaging has advanced significantly with the introduction of transformer-based models~\cite{Selivanov2023Mar}. These models, particularly those leveraging architectures like the Vision Transformer (ViT) and multimodal transformers, have demonstrated superior performance in generating accurate and contextually rich descriptions of medical images~\cite{He2023Feb}. The ability of transformers to capture long-range dependencies and context within images makes them ideal for medical applications where subtle differences can be diagnostically significant~\cite{Nassiri2024Apr}. In the context of the \dsname dataset, the integration of fine-tuning in captioning models can enhance the models' capability to provide detailed and accurate descriptions that can support clinical decision-making and automated reporting.

\subsection{Visual Question Answering}
\gls{vqa} is an emerging research area that combines image understanding with \gls{nlp} to answer questions about images~\cite{Lu2023Dec}. While \gls{vqa} has seen success in general domains, its application in medical imaging remains nascent due to a lack of specialized datasets~\cite{Joshi2024Apr}. The introduction of question-and-answer annotations in our extended dataset \dsname addresses this gap, providing a rich resource for training \gls{vqa} models tailored to medical diagnostics.

\newpage

\Gls{vqa} in medical imaging has benefited from transformer-based models, which excel at integrating visual and textual information~\cite{Naseem2022Mar}. Models like BERT and its variants and successors, well-established in \gls{nlp}, have been adapted for multimodal tasks including VQA~\cite{Khare,Seenivasan2022Sep}. These models effectively understand and generate responses to complex medical questions by synthesizing information from images and text~\cite{Lin2023Sep}.  The \dsname dataset, with its rich annotations, provides a critical resource for training these models, enabling the development of \gls{ai} systems that can deliver accurate and context-aware responses to medical queries, thereby enhancing diagnostic accuracy and efficiency.

\subsection{Synthetic Medical Image Generation}
Recent advancements in synthetic medical image generation have been significantly driven by diffusion-based models, particularly stable diffusion techniques~\cite{Kazerouni2023Aug,Pan2023May}. These models, a class of generative models, iteratively refine images starting from noise, resulting in highly realistic synthetic outputs~\cite{Pan2023May}. 

The \dsname dataset can be used to train models for text-to-image synthesis to generate synthetic medical images that are not only visually accurate but also reflective of the wide range of conditions seen in real-world clinical settings. These synthetic images can augment real-world data, address issues of data scarcity, and enhance the training of \gls{ai} systems for rare or complex conditions, thereby improving diagnostic tools and educational resources~\cite{Ibrahim2024Jun}.

\section{The \dsname Dataset}\label{sec:dataset}
The \dsname dataset we present in this work is an extension of the publicly available HyperKvasir~\cite{Borgli2020Aug} and Kvasir Instrument~\cite{Jha2021Jan} datasets. This extended dataset incorporates question-and-answer ground truth data, developed in collaboration with medical experts. It covers the entire \gls{gi} tract, including both normal and abnormal findings, as well as images of various surgical instruments used in \gls{gi} procedures, such as colonoscopies and gastroscopies.

\subsection{Dataset Sources}

The visual components of the \dsname dataset are sourced from the HyperKvasir and Kvasir-Instrument datasets. Table~\ref{tab:source-images} presents an overview of the source image categories used in the curation of the dataset. Each of those categories contains images from either normal cases, cases with significant diseases (polyps, esophagitis, and ulcerative colitis), or cases involving medical instruments.

\begin{table}[H]
    \centering     
\begin{tabular}{|l|c|c|}
    \hline
    \textbf{Image Category} & \textbf{Number of Samples} & \textbf{Source Dataset} \\ \hline
    Normal & 2500 & HyperKvasir \\ \hline
    Polyps & 1000 & HyperKvasir \\ \hline
    Esophagitis & 1000 & HyperKvasir \\ \hline
    Ulcerative Colitis & 1000 & HyperKvasir \\ \hline
    Instrument & 1000 & Kvasir-Instrument \\ \hline
    \hhline{--~}
    TOTAL & 6500 & \multicolumn{1}{c}{}\\ \hhline{--~}   
\end{tabular}

    \caption{Findings, number of images and source dataset.}
    \label{tab:source-images}   
\end{table}

\subsection{Annotation Process}

Additional question-and-answer ground truth data were collected with input from medical professionals experienced in \gls{gi} disease diagnostics. Initial annotations were conducted by computer scientists using LabelBox~\cite{labelbox}, followed by verification by medical experts\footnote{Due to time constraints, not all samples were validated by medical experts. A more comprehensive version of the dataset, featuring complete expert verification, is planned for release in the near future.}.

For collecting the ground truth, six types of questions were answered for each image in the dataset, namely: Yes/No questions, single-choice questions, multiple-choice questions, color-related questions, location-related questions, and numerical count questions. Table~\ref{tab:question-types} presents an overview of the question types used for the annotations, along with the questions and answer options provided for each type. The annotations cover various \gls{gi} aspects, including findings (questions 1, 2, 5, 6, 8, 17, 18), abnormalities (questions 9, 12, 15), anatomical landmarks (questions 10, 13, 16), and instruments (questions 11, 14, 19), as well as multimedia aspects, including image artifacts (question 3) and text (question 4).

\begin{table*}
    \centering
    \begin{tabular}{|c|c|p{6.5cm}|p{7cm}|}
    \hline
    \textbf{No.} & \textbf{Type} & \textbf{Question} & \multicolumn{1}{c|}{\textbf{Answer Options}} \\ \hline
    
    1 & \multirow{5}{*}{Yes/No} & Have all polyps been removed? & \multirow{5}{*}{No / Yes / Not relevant} \\ 
    2 &  & Is this finding easy to detect? & \\ 
    3 &  & Is there a green/black box artefact? & \\ 
    4 &  & Is there text? & \\ 
    5 &  & Does this image contain any finding? & \\ \hline
    
    6 & \multirow{3}{*}{Choice (Single)} & What type of polyp is present? & Paris Ip / Paris IIa / Paris Is \\ 
    7 &  & What type of procedure is the image taken from? & Capsule Endoscopy / Colonoscopy / Gastroscopy \\ 
    8 &  & What is the size of the polyp? & <5mm / 5-10mm / 11-20mm / >20mm \\ \hline
    
    9 & \multirow{3}{*}{Choice (Multiple)} & Are there any abnormalities in the image? & Oesophagitis, Ulcerative Colitis, Short-Segment Barretts, Barretts, Polyp, Hemorrhoids \\ 
    10 &  & Are there any anatomical landmarks in the image? & Ileum, Z-Line, Cecum, Pylorus \\ 
    11 &  & Are there any instruments in the image? & Tube, Metal Clip, Polyp Snare, Injection Needle, Biopsy Forceps \\ \hline
    
    12 & \multirow{2}{*}{Choice (Color)} & What color is the abnormality? & landmark:grey, flesh, pink, black, orange, etc. \\ 
    13 &  & What color is the anatomical landmark? & pink, red, etc. \\ \hline
    
    14 & \multirow{3}{*}{Location} & Where in the image is the instrument? & Upper-Left / Upper-Center / Upper-Right / \\ 
    15 &  & Where in the image is the abnormality? & Center-Left / Center / Center-Right / Lower-Left / \\ 
    16 &  & Where in the image is the anatomical landmark? & Lower-Center / Lower-Right \\ \hline
    
    17 & \multirow{3}{*}{Numerical Count} & How many findings are present? & \multirow{3}{*}{[0-inf]} \\ 
    18 &  & How many polyps are in the image? & \\ 
    19 &  & How many instruments are in the image? & \\ \hline

\end{tabular}

    \caption{Question types with the question and answer options used for the annotations.}
    \label{tab:question-types}
\end{table*}

\subsection{Final Dataset}

The \dsname dataset comprises 6,500 images, each annotated with various question-and-answer pairs. This dataset is fully accessible on HuggingFace. A subset of the \dsname dataset, comprising 2,000 images and 20,241 captions derived from the annotations for those images, was recently used in a multimedia retrieval challenge~\cite{hicks2023overview}, which includes two specific tasks:

\begin{enumerate}

\item \textbf{\gls{task-is}}: This task involves participants using text-to-image generative models to create a diverse dataset of medical images based on textual prompts. For instance, participants might receive prompts such as "An early-stage colorectal polyp" and are expected to generate an image that accurately represents this description. The development dataset provided includes prompt and image pairs to aid in developing solutions. During the testing phase, participants will be given a list of prompts and must generate one image per prompt to submit to the organizers.

\item \textbf{\gls{task-opg}}: This task requires participants to generate images using self-created prompts within defined categories. Examples include generating images with a specific number of polyps, a polyp in a designated area of the image, or a polyp of a particular type and size. Other categories may involve creating images without findings in the esophagus or large bowel, or including specific instruments such as biopsy forceps, metal clips, or tubes. Additionally, participants might need to create images featuring anatomical landmarks like the Z-line, pylorus, or cecum. The evaluation will consider the quality of the synthetic images, the complexity of the models and prompts, and the hardware requirements. 
\end{enumerate}

In this work, we chose to use the same subset of 2,000 images across the experiments to showcase different applications of the dataset. It is also worth noting that the images in the subset, along with their annotations, were randomly selected in equal proportions from the four different image categories (excluding the "Normal" category) as listed in Table~\ref{tab:source-images}.

The existing annotations for the images were then converted to single-sentence captions using a script. The script generated a templatized caption for each of the annotations. 
The 20,241 textual captions generated from the annotations in the subset are used in both the image captioning and image generation tasks described below.

A large language model (LLM), LLaMA-3 (7B)\footref{llama3}, has been employed to transform these captions into question-and-answer pairs, making them suitable for the VQA task as well. The model was guided by a carefully designed prompt with explicit instructions, and the output was subjected to a retry mechanism and manual review by domain experts to ensure quality and relevance.
The process is explained in detail in the following sections.

\section{Experiments}\label{sec:experiments}

In this section, we present three preliminary experiments to demonstrate the effectiveness and applicability of the \dsname dataset: Image Captioning, \acrfull{vqa}, and Synthetic Medical Image Generation. These tasks were selected to highlight different aspects of the dataset and its potential use in various medical imaging and diagnostic contexts. Our implementation for all experiments is available open source.

The results from our experiments underscore the versatility and robustness of the \dsname dataset, providing a foundation for future advancements in medical image analysis and diagnostics. The results indicate that models trained on \dsname can effectively perform various vision-language tasks, showcasing the dataset's potential in real-world medical applications.

\subsection{Image Captioning}

\hspace{\parindent}\textbf{Model and Setup:} For the image captioning task, we used Florence-2~\cite{Xiao2023Nov}, an open-source, lightweight vision-language model, known for its strong zero-shot and fine-tuning capabilities across various tasks, including captioning, object detection, grounding, and segmentation. We fine-tuned Florence-2 with the prefix {<DETAILED\_CAPTION>}. The model consists of 0.23 billion parameters, and we applied \gls{lora}~\cite{Hu2021Jun} for efficient fine-tuning.

To optimize training efficiency and resource usage, we froze the image encoder during fine-tuning. This decision was based on the observation that while unfreezing the image encoder could potentially enhance performance, it would also significantly increase computational demands and resource consumption. By maintaining the image encoder in a frozen state, we focused the training process on the language model component, allowing us to achieve substantial improvements in caption generation with reduced resource requirements.

During training, we used a learning rate of 1.8e-6 and employed a linear scheduler to manage the learning rate over the course of training. The model was trained for ten epochs with a batch size of 20 on NVIDIA A100 GPU, utilizing the AdamW optimizer to update model weights. We monitored training and validation loss to assess model performance and adjust training parameters as needed. The model and processor were saved at each epoch, facilitating incremental evaluation and potential model deployment.

\textbf{Dataset and Training:} A subset of \dsname was employed for this task, where each medical image is paired with a descriptive caption. The whole of the data-subset with 2,000 images was used for the training. The model was trained for 10 epochs, ensuring the loss had stabilized. During inference, the model generates the top 5 candidate captions for each image, acknowledging that different valid aspects of the image might be highlighted.

\begin{figure}[ht]
\includegraphics[width=0.5\columnwidth]{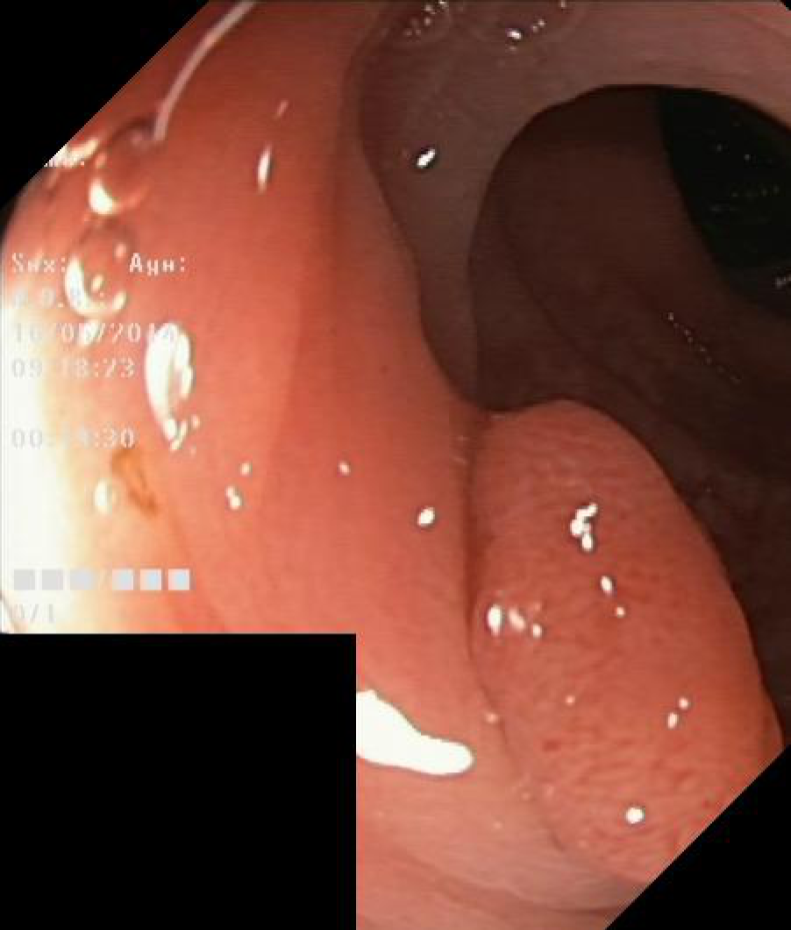}
\small\\
Captions generated by the model (5 generations):\\
    \begin{enumerate}
    \item a polyp of type paris iia
    \item a polyp of type with the color red
    \item a polyp of type paris
    \item a green/black box artefact
    \item a polyp image from a colonoscopy
\end{enumerate}
    \caption{[Task 1] An example from the fine-tuned captioning model, generating five captions for the given input image.}
    \Description[Example of model-generated captions]{The image shows an example from a fine-tuned captioning model. It displays five different captions generated for a given input image. Each caption describes various aspects of the image, providing insights into how the model interprets visual content.}

    \label{fig:example-captioning}
\end{figure}

\textbf{Evaluation Metrics:} We evaluated the fine-tuned model using standard captioning metrics such as \gls{bleu}, \gls{meteor}, and \gls{cider}~\cite{Vedantam} over 5,000 image-caption pairs from the data-subset.
The results are presented in Table~\ref{tab:evaluation-metrics}, demonstrating the model's capability to accurately describe medical images. An example of caption generation\footnote{\label{AIwarning}The results are generated by an \gls{ai} model and may not always accurately reflect true medical conditions or diagnoses.} from the trained model is presented in Figure~\ref{fig:example-captioning}.

\begin{table}[H]

    \centering
    \begin{tabular}{|c|c|c|}
    
        \hline
        \textbf{Task}
        & \textbf{Metric}
        & \textbf{Score} 
        \\ \hline
        
        \multirow{3}{*}{Image Captioning}
        & BLEU & 0.0823 \\ 
        & ROUGE & 0.3905 \\
        & METEOR & 0.2632 \\
        & CIDEr & 0.2642 \\
        \hline

        & BLEU &  0.3757 \\ 
        Visual Question& ROUGE & 0.6955 \\
        Answering& METEOR & 0.6640 \\
        & CIDEr & 0.7320 \\
        \hline
        Synthetic Medical & FID & 110.73 \\ 
        Image Generation & IS & 3.07; 0.14 \\
         & IS (real) & 4.05; 0.24 \\
        \hline 

    \end{tabular}
    \caption{Model evaluation metrics for the three tasks. \small{FID is Fréchet Inception Distance, and IS is Inception Score. IS (real) represents the Inception Score for the real image dataset used in training. The average and standard deviation of the Inception Score are reported.} }
    \label{tab:evaluation-metrics}
    
\end{table}

\subsection{Visual Question Answering}

\textbf{Synthetic \gls{vqa} Dataset Generation} We generated a synthetic \gls{vqa} dataset by leveraging the capabilities of the LLaMA-3 (7B)\footnote{\label{llama3}\href{https://github.com/meta-llama/llama3}{https://github.com/meta-llama/llama3}} language model. The primary objective was to create diverse and contextually relevant question-and-answer pairs from existing image captions, thereby producing a valuable resource for training and evaluating \gls{vqa} systems.

We commenced with the curated set of image captions. These captions already provided descriptive information about the images, encompassing visible properties, procedural details, and notable findings.

\begin{itemize}
\item Language Model Selection:
The LLaMA-3 (7B) model, a state-of-the-art language model known for its adeptness in understanding and generating human-like text, was selected for this task. The model's large parameter size and advanced training regimen equipped it to handle the complex and domain-specific language inherent in medical captions.

\item Prompt Design:
To guide the language model in generating question-and-answer pairs, we designed a comprehensive prompt, which included:
\begin{itemize}
    \item A role specification for the model, indicating its function as an intelligent dataset generator.
    \item Explicit instructions to formulate questions based on the content of the captions, ensuring that the questions did not refer directly to the images. This was crucial for maintaining generalizability and focusing on the textual information.
    \item Specific guidelines for generating questions related to aspects such as the presence of text, the type of procedure, polyp count, color, size, and location, when mentioned in the caption.
\end{itemize}

\item Data Generation Process:
Utilizing the LLaMA-3 (7B) model, for each caption, a JSON object was generated containing the question (Q) and answer (A) pairs. This process involved:
\begin{itemize}
    \item Parsing the caption to identify key elements.
    \item Crafting a question that emphasizes these elements.
    \item Formulating an answer that accurately reflects the information conveyed in the caption.
\end{itemize}

\item Quality Control:
To ensure the quality and relevance of the generated data, several quality control measures were implemented:
\begin{itemize}
    \item Retry Mechanism: The data generation process included a retry mechanism to handle instances where the output was incomplete or non-informative, thereby ensuring the robustness of the dataset.
    \item Manual Review: A subset of the generated question-and-answer pairs was subjected to manual review by domain experts to evaluate the appropriateness of the questions and the correctness of the answers.
\end{itemize}

\item Output and Storage:
The resulting question-and-answer pairs were saved, organized according to the index of the original captions. This systematic organization facilitates easy access and integration into VQA systems for training and evaluation.
\end{itemize}

\hspace{\parindent}\textbf{Model and Setup:} To address the limitations of image captioning in controlling specific aspects of image information, we leveraged the \gls{vqa} task. We fine-tuned the same Florence-2 model with the prefix \texttt{<MedVQA>} and included both the input image and the corresponding question in the input. This task benefits from easier evaluation compared to captioning, as the model's output is constrained to answering specific questions about the image.

\textbf{Dataset and Training:} We used the synthetic \gls{vqa} dataset generated by prompting the LLaMA-3 (7B) language model\footref{llama3} to create relevant question-and-answer pairs from the existing captions. Similar to the image captioning task, the subset of \gls{vqa} dataset with 2,000 images was used in the training for synthetic VQA generation, and 20,241 question-and-answer pairs were  synthesized from then captions. The model was fine-tuned for 10 epochs. 

\begin{figure}[ht]

\includegraphics[width=0.7\columnwidth]{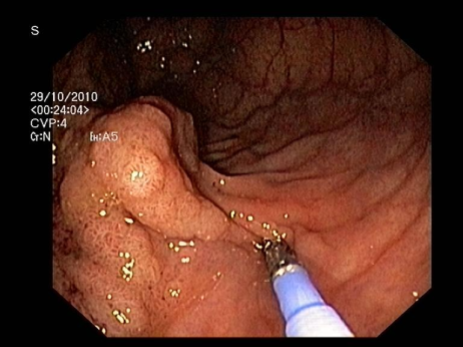}

\small
\begin{enumerate}
    \item \textbf{Q:} How many polyps are seen? \\
          \textbf{A:} 1 polyp is seen.
    \item \textbf{Q:} What is the color? \\
          \textbf{A:} It's pink.
    \item \textbf{Q:} Are there any medical instruments? \\
          \textbf{A:} Yes, there is one.
\end{enumerate}

\caption{[Task 2] An example from the fine-tuned \gls{vqa} model answering questions about the input image.}
\Description[Example of VQA model oputput]{This image shows a VQA (Visual Question Answering) model in action. It displays an input image with several questions posed about the content. The model's answers to these questions are shown alongside the image. The image helps illustrate how the model interprets and responds to visual information.}

    \label{fig:example-vqa}
\end{figure}

\textbf{Evaluation Metrics:} We assessed the performance of the \gls{vqa} model using metrics such as BLEU, ROUGE, METEOR, and CIDEr on all the question-and-answer pairs generated from the subset. The detailed results are shown in Table~\ref{tab:evaluation-metrics}, highlighting the model's proficiency in answering specific medical questions based on image content. An example\footref{AIwarning} of question answering from the trained model is presented in Figure~\ref{fig:example-vqa}.

\subsection{Synthetic Medical Image Generation}

\hspace{\parindent}\textbf{Model and Setup:} To generate high-quality synthetic medical images, we trained the Stable Diffusion 3 model~\cite{Esser2024Mar}, which utilizes an advanced \gls{mmdit} architecture. This model translates textual prompts into high-resolution images, thereby enhancing the dataset's utility for training and testing various models. The training employed the DreamBooth technique~\cite{DreamBooth}, which updates the entire diffusion model by training on just a few images of a subject or style, associating a special word in the prompt with the example images. A resolution of 512x512 pixels was used, and a large batch size of 48 was employed with gradient accumulation steps set to 1 on the NVIDIA A100 GPU. Mixed precision (fp16) training and the Prodigy optimizer were utilized~\cite{Mishchenko2023Jun}, alongside a sigma-sqrt weighting scheme and a constant learning rate. The model underwent validation at every epoch to evaluate the generated output with set prompts, and gradient checkpointing was enabled to optimize memory usage.

\textbf{Dataset and Training:} The input prompts for the synthetic medical image generation task were derived from the captions used in the image captioning task. We focused on generating high-quality synthetic images that accurately represent the textual descriptions. The model's fine-tuning process included the application of a low-rank adaptation (\gls{lora}) with a rank value of 128, allowing for efficient parameter updates without overfitting. 
The model, containing 2 billion parameters, was fine-tuned using \gls{lora} for 80 epochs until the training loss had stabilized.
Post-training evaluation demonstrated that the model could consistently produce anatomically plausible and diagnostically relevant images, making it a valuable tool for generating medical image datasets.

\begin{figure}[ht]
    \small
    \textbf{Generation Prompt}: a non polyp image with an instrument
    \includegraphics[width=1\columnwidth]{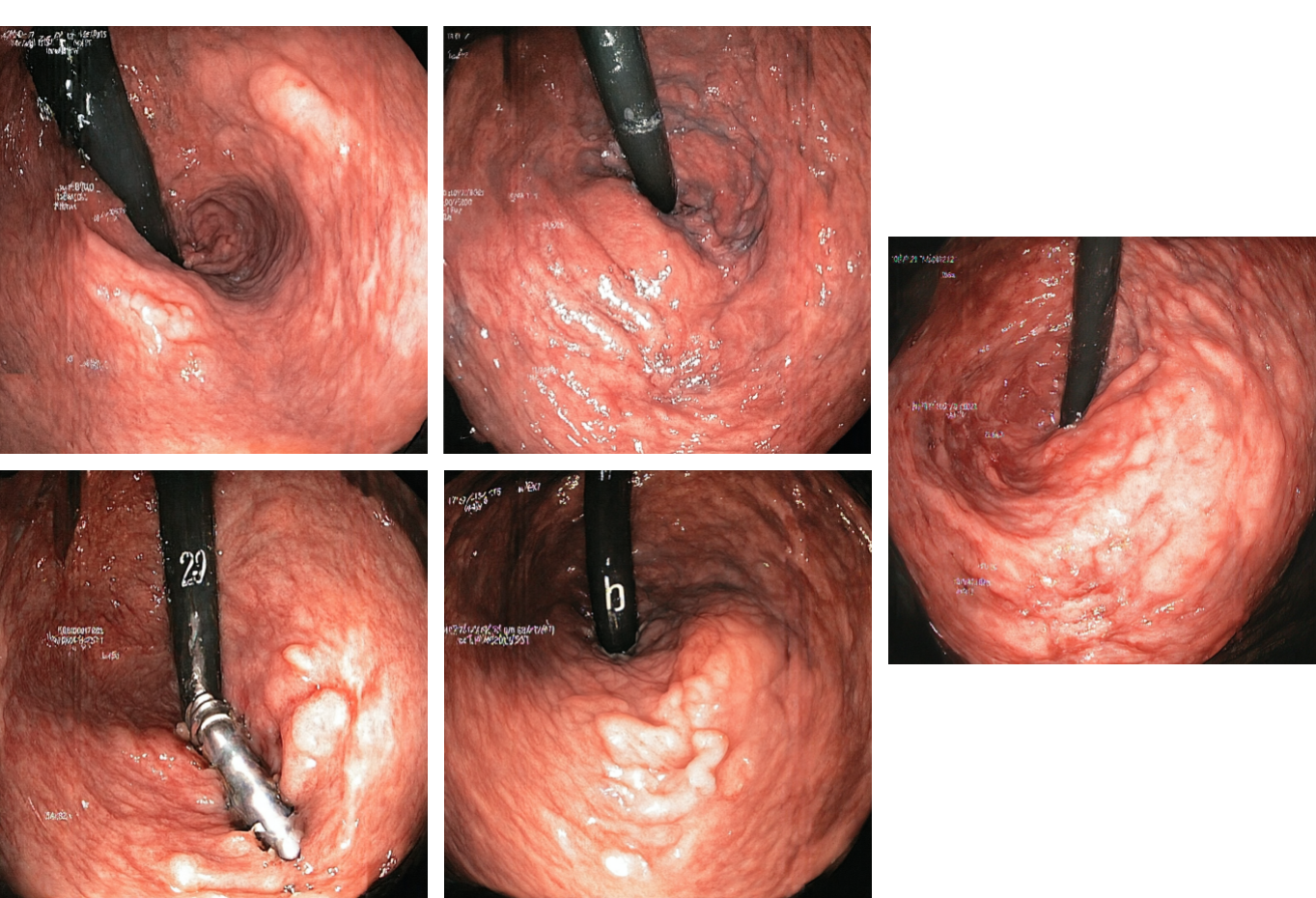}
    \caption{[Task 3] An example from the fine-tuned synthetic medical image generation model, which generated five images for the given prompt.}
  \Description[Examples of generated medical images]{This figure contains five synthetic medical images generated by a fine-tuned model. The images depict the model's ability to produce diverse and realistic medical visuals based on the provided prompt.}

    \label{fig:example-imagen}
\end{figure}

\textbf{Evaluation Metrics:} We first generated 5,000 images each for the polyp and non-polyp classes using randomly selected prompts from the data subset. The generated images in these two classes were then evaluated using metrics such as \gls{fid} and \gls{is} to assess the quality and diversity of the synthetic images. The quality and diversity of the real images in these two classes from the subset were also calculated for comparison using the \gls{is} score. 
Table~\ref{tab:evaluation-metrics} presents the evaluation results, demonstrating the effectiveness of the model in generating realistic medical images.
The \gls{fid} score and \gls{is}  for the generated images indicate that while the model produces images with some level of realism, there is still a gap compared to real images. A lower FID score and a higher IS value generally correspond to more realistic and diverse images. Therefore, the results suggest that the generated images, while somewhat effective, still have room for improvement in terms of visual quality and diversity. An example of image generation\footref{AIwarning} from the trained model is presented in Figure~\ref{fig:example-imagen}.

\section{Discussion}\label{sec:discussion}

The \dsname dataset aims to bridge the gap between medical image analysis and practical diagnostic applications, ultimately contributing to improved patient care and diagnostic accuracy. Below, we elaborate on dataset applications, as well as limitations and potential future work.

\subsection{Dataset Applications}\label{sec:applications}
The \dsname dataset extends the capabilities of existing \gls{gi} datasets by introducing comprehensive question-and-answer annotations for medical images. This enhancement allows for the development of sophisticated \gls{ml} models that can perform a variety of tasks essential for medical diagnostics.

\begin{itemize}

    \item \textbf{Image Captioning}: As demonstrated in Section~\ref{sec:experiments}, \dsname can be used to train captioning models. By generating descriptive captions for medical images, it is possible to automate the creation of detailed medical reports, reducing the burden on healthcare professionals and minimizing the risk of human error. The Florence-2 model, fine-tuned for the image captioning task, has shown impressive performance in generating accurate and relevant captions, as evidenced by high \gls{bleu}, \gls{meteor}, and \gls{cider} scores.

    \item \textbf{\acrfull{vqa})}: The incorporation of \gls{vqa} tasks enables models to interpret medical images in a more interactive and detailed manner. Our experiments with the Florence-2 model, fine-tuned for \gls{vqa}, demonstrate the model's ability to accurately answer specific medical questions based on visual input. This capability is crucial for developing diagnostic tools that can assist clinicians by providing immediate, context-specific information about \gls{gi} conditions. The ability for the model to provide text as output might also enhance healthcare professionals' interaction with the model because the prediction includes more context than just the predicted class or diagnosis. This can make it easier to `quality check' the model´s reasoning behind the prediction and contribute to a more natural way of interacting with a machine. 

    \item \textbf{Synthetic Medical Image Generation}: The ability to generate high-quality synthetic medical images using models such as Stable Diffusion 3 opens new avenues for data augmentation and training. Synthetic images can help address the issue of class imbalance in medical datasets and provide additional training samples for rare conditions. Our experiments demonstrate the effectiveness of \dsname for synthetic medical image generation, with generated images exhibiting high fidelity and diversity as measured by \gls{fid} and \gls{is} metrics.
    
    \item \textbf{Object Detection and Localization}: The question-and-answer annotations related to the location of abnormalities, instruments, and anatomical landmarks enable precise training of object detection and localization models. These models are essential for tasks such as polyp detection and surgical instrument recognition, which are critical for real-time diagnostic support during endoscopic procedures.
    
    \item \textbf{Classification}: The yes/no and choice questions in the \dsname dataset facilitate the development of classification models. These models can assist in diagnosing specific \gls{gi} conditions, identifying procedural contexts, and recognizing the presence of surgical instruments. \dsname supports both single-choice and multiple-choice questions, allowing for a comprehensive evaluation of classification model performance. 
    
\end{itemize}

\subsection{Synthetic VQA Dataset Generation}
The synthetic VQA dataset generated with LLM demonstrated a broad vocabulary and diverse question structures, showcasing the model’s ability to understand and interpret complex medical language. This linguistic variety is crucial for developing robust VQA systems that can handle a wide range of questions and scenarios in medical imaging.

Using the LLaMA-3 (7B) model to generate synthetic VQA datasets has proven effective, offering a scalable and efficient method for producing large datasets with varied linguistic patterns. This approach not only improves training data for VQA systems but also advances medical \gls{ai} by fostering the development of more nuanced models. Future work could refine prompt engineering and integrate iterative feedback to further enhance dataset quality.

\section{Limitations and Future Work}\label{sec:limitations}

While the \dsname dataset marks significant progress, it is crucial to acknowledge that several limitations remain. Addressing these limitations presents opportunities for meaningful improvements and further advancements in the dataset’s utility and accuracy.

\begin{itemize}
    \item \textbf{Expert Verification of Annotations:} As outlined in Section~\ref{sec:dataset}, the initial annotations in the \dsname dataset were performed by computer scientists, with subsequent verification by medical professionals. However, due to time constraints, not all annotations received expert validation. Future work will include the release of an enhanced version of the dataset, which will undergo complete expert verification to ensure the highest standards of annotation accuracy and reliability.

    \item \textbf{Scope:} The current \dsname dataset includes samples from a range of \gls{gi} conditions and procedural contexts. However, it does not cover the full spectrum of GI conditions encountered in clinical practice. Future efforts will focus on expanding the dataset to encompass a wider array of GI conditions and procedural contexts, thereby increasing its comprehensiveness and applicability for diverse diagnostic applications.

    \item \textbf{Scale:} The \dsname dataset currently comprises 6,500 annotated images, a subset of the images available in the source datasets, such as HyperKvasir, which contains over 10,662 labeled images. 
    
    We aim to expand the dataset by integrating additional images and annotations from these source datasets, thereby providing a more extensive resource for training and evaluating machine learning models.

    \item \textbf{Validation:} All experiments and results reported in this study are based on a subset of 2,000 images, including their corresponding annotations, derived captions, and question-and-answer pairs. For reproducibility, comparative analysis, and proper validation of the trained models, the full dataset should be utilized with appropriate training, validation, and test data splits. Addressing this will be considered in future work.

    \item \textbf{Development of Diagnostic Tools:} While the \dsname dataset is designed to support the development of AI-driven diagnostic tools, the current iteration may not fully address all necessary components for comprehensive diagnostic support. Future enhancements will aim to align the dataset with emerging clinical requirements and technological advancements, thereby more effectively contributing to the development and refinement of diagnostic tools and systems in healthcare.

\end{itemize}

Future work will prioritize the augmentation of the dataset through additional expert-verified annotations and an expanded scope that includes a broader range of GI conditions and procedural contexts.
By continuously improving and broadening this resource, we aspire to facilitate the development of advanced AI-driven diagnostic tools that have the potential to transform medical imaging and diagnostics.

\section{Conclusion}\label{sec:conclusion}

In this paper, we introduce the \dsname dataset, an extension of the existing HyperKvasir and Kvasir-Instrument datasets, enriched with question-and-answer annotations. This dataset is specifically designed to advance research in medical image analysis, particularly in the field of gastrointestinal diagnostics. The \dsname dataset facilitates a variety of applications, including image captioning, VQA, and synthetic medical image generation, as demonstrated through our preliminary experiments.

The introduction of synthetic question-and-answer pairs provides a new dimension to the dataset, enabling the development of sophisticated \gls{ai} models capable of nuanced understanding and interactive diagnostics. These capabilities are crucial for improving the accuracy and efficiency of medical diagnostics, thereby improving patient care.
Despite its promising potential, the \dsname dataset is still a work in progress. Future efforts will focus on extending the dataset’s scope and scale, ensuring comprehensive expert validation of annotations, and aligning with clinical needs. These enhancements will further solidify the dataset's role as a valuable resource for the medical community, supporting the development of next-generation diagnostic tools.

\begin{acks}
This work has benefited from the Experimental Infrastructure for Exploration of Exascale Computing (eX3), which is financially supported by the Research Council of Norway under contract 270053.
\end{acks}

\bibliographystyle{ACM-Reference-Format}
\balance
\bibliography{references}

\end{document}